\documentclass{article}
\usepackage{ijcai22}

\usepackage{times}
\usepackage{soul}
\usepackage{url}
\usepackage[hidelinks]{hyperref}
\usepackage[utf8]{inputenc}
\usepackage[small]{caption}
\usepackage{graphicx}
\usepackage{amsmath}
\usepackage{amsthm}
\usepackage{booktabs}
\usepackage{algorithm}
\usepackage{algorithmic}
\usepackage{amssymb}
\usepackage{multirow}
\title{Subgraph Neighboring Relations Infomax for Inductive Link Prediction on Knowledge Graphs}

\author{
Xiaohan Xu$^{1,2}$\and
Peng Zhang$^1$\footnote{Contact Author}\and
Yongquan He$^{1,2}$\and
Chengpeng Chao$^{1,2}$\And
Chaoyang Yan$^{1,2}$\\
\affiliations
$^1$Institute of Information Engineering, Chinese Academy of Sciences\\
$^2$School of Cyber Security, University of Chinese Academy of Sciences\\
\emails
\{xuxiaohan, pengzhang, heyongquan, chaochengpeng, yanchaoyang\}@iie.ac.cn
}

\begin{document}

\maketitle

\begin{abstract}
Inductive link prediction for knowledge graph aims at predicting missing links between unseen entities, those not shown in training stage. Most previous works learn entity-specific embeddings of entities, which cannot handle unseen entities. Recent several methods utilize enclosing subgraph to obtain inductive ability. However, all these works only consider the enclosing part of subgraph without complete neighboring relations, which leads to the issue that partial neighboring relations are neglected, and sparse subgraphs are hard to be handled. To address that, we propose Subgraph Neighboring Relations Infomax, SNRI, which sufficiently exploits complete neighboring relations from two aspects: \textit{neighboring relational feature} for node feature and \textit{neighboring relational path} for sparse subgraph. To further model neighboring relations in a global way, we innovatively apply mutual information (MI) maximization for knowledge graph. Experiments show that SNRI outperforms existing state-of-art methods by a large margin on inductive link prediction task, and verify the effectiveness of exploring complete neighboring relations in a global way to characterize node features and reason on sparse subgraphs. \footnote{Code and data are available at \url{https://github.com/Tebmer/SNRI}}
\end{abstract}

\section{Introduction}
Knowledge graphs (KGs) are collections of structured knowledge represented by factual triples (\textit{entity, relation, entity}), which are essential for many applications, such as question answering \cite{huang2019kgqa}, recommendation systems \cite{wang2018ripple}. However, even state-of-the-art KGs suffer from incompleteness issue, e.g. FreeBase \cite{boll2008freebase}, and WikiData \cite{vran2012wiki}. To complete KGs, link prediction task aims at inferring missing links between entities on original KGs. But in fact, there are many newly emerging entities added into real-world KGs constantly over time \cite{Trivedi2017ke}, e.g., new user added into e-commerce database or new molecules in biomedical KGs. In order to predict links between brand-new entities, inductive link prediction task has been an active area of research, which requires model with the inductive ability for reasoning on graphs consisting of unseen nodes.

\begin{figure}[t]
  \centering
  \includegraphics[scale=0.36]{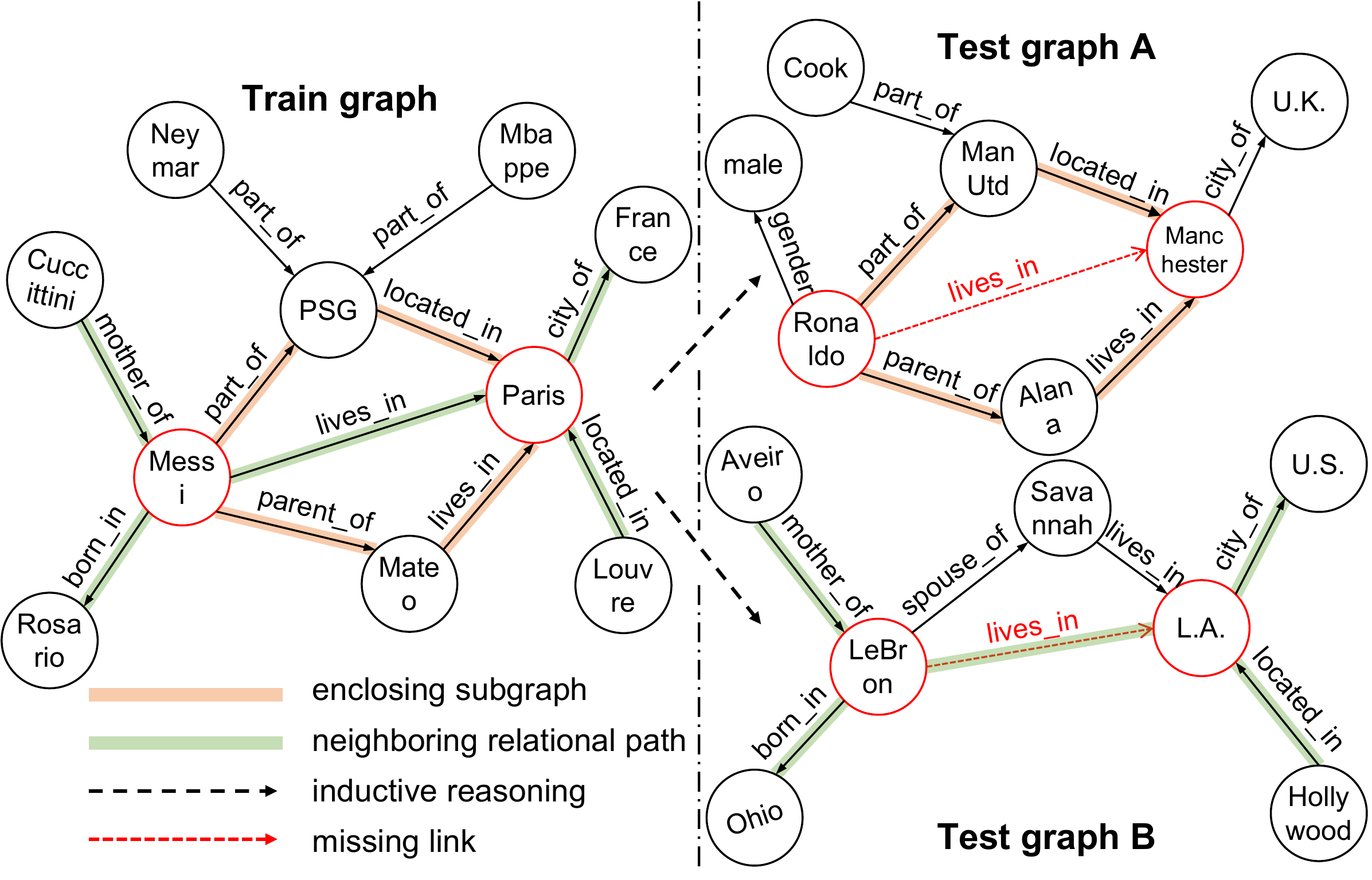}
  \caption{Two explanatory cases in inductive link prediction. Methods based on enclosing subgraph (red paths) can reason on Test graph A, but hard to handle sparse Test graph B. In contrast, our work utilizes neighboring relations not included in enclosing subgraph simultaneously to build neighboring relational paths (green paths) for reasoning on sparse Test graph B. 
    }\label{fig:eg}
\end{figure}

Whereas inductive link prediction is a difficult task as it requires generalization from training entities to unseen entities. Most previous link prediction methods \cite{bordes2013transe,yang2015distmult} learn specific embedding for each entity, which are hard to generalize to unseen entities.
Recently, motivated by graph neural network (GNN) with the ability of aggregating local information, several inductive models based on GNN have been proposed. GraIL \cite{teru2020grail} models enclosing subgraph of target triple to capture topological structure (see Test graph A in Figure \ref{fig:eg}), which owns inductive ability. On the basis of GraIL, several works \cite{Chen2021tact,mai2021compile} further utilize enclosing subgraph structure to predict links inductively. However, all above methods 
only consider the enclosing part of subgraph without complete neighboring relations, which leads to two challenging issues. 
First, they lose partial neighboring relations due to the nature of enclosing subgraph. But all neighboring relations contain valuable information to characterize entities (called \textit{neighboring relational feature}). For example, in Figure \ref{fig:eg}, part of relations \textit{mother\_of} and \textit{born\_in} around node \textit{Messi} are excluded from enclosing subgraph (red paths in Train graph), but they characterize the ``human'' attribution of \textit{Messi}.
Second, enclosing subgraph may be empty or sparse,
and all above methods cannot work well in this case, e.g. no usable connecting path existing between \textit{LeBron} and \textit{L.A.} in Test graph B. In this case, all above methods cannot work without enclosing subgraph. In fact, we can still reason inductively by some relational paths across target nodes (called \textit{neighboring relational path}), e.g. the relational path (\textit{gender}, \textit{lives\_in}, \textit{located\_in}) in Train graph (green paths). 

Based on the above observations, we propose a novel inductive reasoning model, called Neighboring Relational Path Infomax, SNRI, which can effectively exploit complete neighboring information in subgraphs and model neighboring relational paths in a global way by MI maximization. Specifically, SNRI models complete neighboring relations in two aspects: \textit{neighboring relational feature} for node initializing and \textit{neighboring relational path} for sparse subgraph modeling. In contrast to previous works \cite{teru2020grail}, we first extract enclosing subgraph for each triple but reserve complete neighboring relations for each entity. These neighboring relations are then aggregated in an attentive manner to represent entities feature. After that, we utilize neighboring relations of target triples again to build neighboring relational paths in a global way by MI maximization mechanism and apply a joint strategy for training. In this way, SNRI can effectively incorporate complete relational information into enclosing subgraph and model neighboring relational paths, thus improving the performance of inductive link prediction.

Our key contributions are summarized as follows: 
1) We propose a novel inductive reasoning model, SNRI, which effectively integrates complete neighboring relations into the enclosing subgraph from two aspects: neighboring relational feature and neighboring relational path.
2) We innovatively apply MI maximization to inductive link prediction by maximizing local and global representation to model subgraph and neighboring relational paths in a global way.
3) Experiments conducted on benchmark datasets show that our work outperforms existing inductive reasoning by a large margin and demonstrate the effectiveness of characterizing entities and modeling sparse subgraphs.

\section{Related Work}

\subsection{Link Prediction Methods}
\paragraph{Transductive methods.} 
Transductive methods learn an entity-specific embedding for each node, such as 1) translation-based TransE \cite{bordes2013transe} and TransH \cite{Wang2014transh}; 2) factorization-based RESCAL \cite{nickle2012rescal}, and 3) GNN-based R-GCN \cite{sch2018rgcn} and CompGCN \cite{vash2020compgcn}. The major differences among them are the scoring function and whether utilize structure information. However, all above models have one thing in common: reasoning over original KGs, and thus difficult to predict missing links between unseen nodes.

\paragraph{Inductive methods.} 
Inductive models have generalizing ability for reasoning on unseen nodes. They are categorized into rule-based and graph-based methods. Rule-based methods explicitly learn logical rules for reasoning, which is independent to entities and thus inductive. 
Some differentiable methods, NeuralLP \cite{yang2017nlp} and DRUM \cite{sade2019drum}, learn logical rules and rule-confidence simultaneously in an end-to-end differentiable manner. However, they ignore the structure around the target triple, leading to a low expressive ability.
In recent years, graph neural network (GNN) has been a powerful tool in link prediction. Some graph-based methods, such as LAN \cite{wang2019lan}, aggregate neighboring node embeddings to obtain embeddings of unseen nodes, but they have limitation that unseen nodes have to be surrounded by known neighboring nodes. For reasoning inductively by structure information, GraIL \cite{teru2020grail} is the first method proposed to model enclosing subgraph structure around the target triple. 
Inspired by GraIL, CoMPILE \cite{mai2021compile} proposes a communicative message passing network to strengthen the message interactions between edges and entitles, thus enables a sufficient flow of relation information. 
However, all these models based on enclosing subgraph suffer two problems: 1) partial neighboring relations are neglected when extracting the enclosing subgraph, and 2) when enclosing subgraph is sparse or even empty, they are hard to reason inductively. In contrast, our work keeps integrated neighboring relations and builds neighboring relational paths to handle sparse subgraphs, which has a better ability for reasoning.

\begin{figure*}[t]
  \centering
  \includegraphics[scale=0.5]{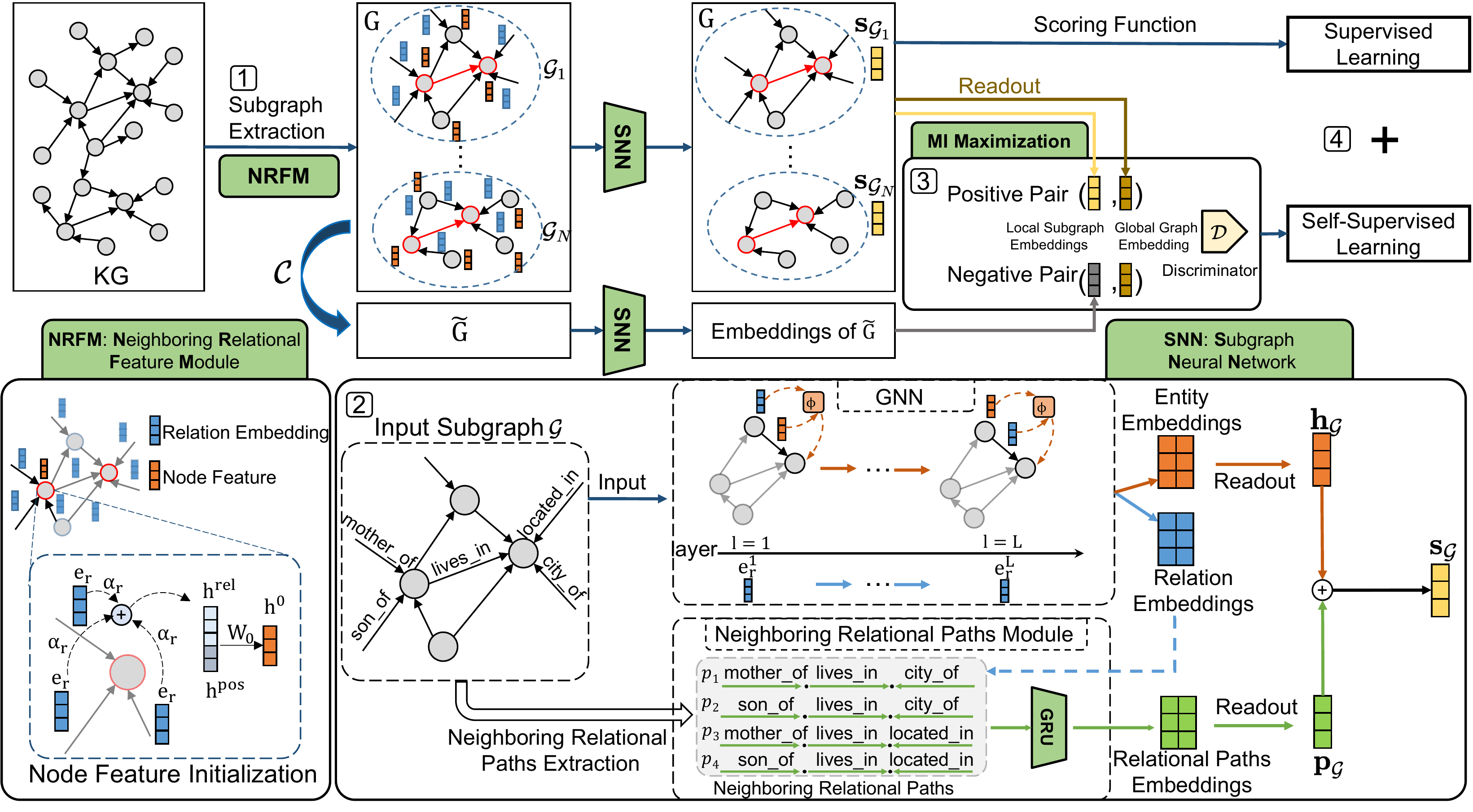}
  \caption{An overview of our proposed SNRI, which consists of the following steps: 1) extract subgraphs with complete neighboring relations, and initialize the node features by neighboring relational features; 2) feed subgraphs into subgraph neural network to learn representations; 3) maximize MI between subgraph-graph to model neighboring relations in a global way, and 4) train model by a joint strategy.
  }\label{fg:framework}
\end{figure*}

\subsection{Contrastive Learning}
Contrastive learning is an important approach of self-supervised learning, which trains an encoder to be contrastive between representations that captures statistical dependencies of interest and those that do not \cite{vel2019dgi}. Contrastive Learning has shown great superiority in many downstream applications \cite{devlin2019bert,he2020mc}. Recently, many works \cite{qiu2020gcc,vel2019dgi} apply contrastive Learning for GNN. DGI \cite{vel2019dgi} maximizes mutual information (MI) between local representation and global representation of graph to capture more common local features from both local and global perspectives. However, DGI only applies MI to unweighted graph with simple relations. Motivated by DGI, DRGI \cite{liang2021drgi} introduces MI to knowledge graphs to handle multi-relational graphs, but it is still a transductive model. For catching neighboring relations in a global way, we innovatively apply MI into the inductive link prediction task by maximizing MI of subgraph and graph representations.

\section{Methods}
In this section, we introduce our proposed method SNRI in detail. 
The overall task is to score a triple $ (u, r_t, v) $ in a KG $ G=|V,R| $ inductively, i.e. to predict the likelihood of the \textit{target relation}  $ r_t $ between the unseen \textit{target nodes} $ u$ and $ v $, where $ V $ and $ R $ are sets of nodes and relations.     
An overview of our proposed SNRI is shown in Figure \ref{fg:framework}.
SNRI mainly consists of four parts: 1) subgraph extraction and neighboring relational feature module to initialize node features, 2) subgraph neural network to learn representations of subgraphs, 3) self-supervised mutual information mechanism to model neighboring relations in a global way, and 4) a joint training strategy to optimize model. 

\subsection{Neighboring Relational Feature Module}

\paragraph{Subgraph extraction.} We first extract enclosing subgraph $ \mathcal{G}(u, r_t, v) $  around target triple $ (u, r_t, v) $ following  GraIL \cite{teru2020grail}. There are three steps for subgraph extraction. First, we obtain node sets of $ k $-hop neighborhood, $ \mathcal{N}_k(u) $ and $ \mathcal{N}_k(v) $, of two target nodes $ u $ and $ v $ respectively. Then, we obtain the enclosing subgraph by taking intersection of $\mathcal{N}_k(u) \cap  \mathcal{N}_k(v) $. In the end, we filter out nodes that are isolated or at a distance greater than $ k $  from either of the target nodes. But different from GraIL, we reserve complete neighboring relations $ \mathcal{N}^r(u) $ of each node, which contains relations partially omitted by enclosing subgraph.

\paragraph{Node initialization.}
Since inductive reasoning demands node attributes cannot be used, and GNN requires a node feature matrix $ \mathbf{X} $ as input \cite{gilmer2017mp}, our work initialize the node feature $ \mathbf{h}^0_i$ for node $ i $ by combining positional feature $ \mathbf{h}^{pos}_i$ and neighboring relational feature $ \mathbf{h}^{rel}_i$ (see lower left of Figure \ref{fig:eg}). First, we obtain the positional feature $ \mathbf{h}^{pos}_i \in \mathbb{R}^{d_p}$ by double radius vertex labeling \cite{zhang2018gnn} scheme :
\begin{equation}
  \mathbf{h}^{pos}_i = [\text{one-hot}(d(i, u)) \oplus \text{one-hot}( d(i, v) ])),
\end{equation}
where $ d(i, u) $ and $ d(i, v) $ denote the shortest distance from node $ i $ to target head node $ u $ and target tail node $ v $; $ \oplus $ represents the concatenation operation; Second, we propose the following message passing for node $ i $ in an attentive manner to capture neighboring relational feature $ \mathbf{h}^{rel}_i \in \mathbb{R}^{d}$ :
\begin{align}
  \mathbf{h}^{rel}_i  &= \sum_{r \in \mathcal{N}^r(i)} \alpha_r \mathbf{e}_r, \\
  \alpha_r = \text{softmax}(\mathbf{e}_r, \mathbf{e}_{r_t}) &= \frac{\text{exp}(\mathbf{e}_r^\top \mathbf{e}_{r_t})}{\sum_{r^\prime \in \mathcal{N}^r(i)}\text{exp}(\mathbf{e}_{r^\prime}^\top \mathbf{e}_{r_t})},
\end{align}
where $ \mathbf{e}_r $ and $ \mathbf{e}_{r_t} $ are relation embeddings of neighboring relation $ r $ and target relation $ r_t $, and $ \alpha_r $ reflects the importance of relation $ r $ to node $ i $ under target relation $ r_t $.
In the end, we represent feature $ \mathbf{h}_i \in \mathbb{R}^d $ of node $ i $ by concatenation of $\mathbf{h}^{rel}_i $ and $\mathbf{h}^{pos}_i$, and project node embeddings to the same embeddings space as relations by $ \mathbf{W}_0 \in \mathbb{R}^{(d+d_p) \times d } $ :
\begin{equation}
  \mathbf{h}^0  _i = \mathbf{W}_0 [\mathbf{h}^{rel}_i \oplus \mathbf{h}^{pos}_i].
\end{equation}
We argue that the feature of nodes with complete neighboring relational semantics are more expressive and robust.


\subsection{Subgraph Neural Network}

With the initial feature of nodes, we input sampled subgraphs to subgraph neural network in SNRI (see lower right of Figure \ref{fg:framework}). As the main component of SNRI, the subgraph neural network models subgraph by two steps: 1) obtain representation of enclosing subgraph by GNN; 2) extract and model neighboring relational paths across target triple.

\subsubsection{Enclosing Subgraph Module}

We first input the subgraph $ \mathcal{G}(u, r_t, v) $ of target triple $ (u, r_t, v) $ to GNN to learn representation of enclosing subgraph. For sufficiently modeling correlations between relations, our GNN model considers the interaction between nodes and relations. We define our nodes' updating function in $ k $-th layer as: 
\begin{equation}
    \mathbf{h}_{i}^{k} =\sum_{r \in R} \sum_{j \in \mathcal{N}_{r}(i)} \alpha_{i, r} \mathbf{W}_{r}^{k} \phi(\mathbf{e}^{k-1}_r, \mathbf{h}_{j}^{k-1}),
\end{equation}
\begin{equation}
  \alpha_{i, r} =\sigma_{2}\left(\mathbf{W}_{2} \boldsymbol{c}_{i, r}+\boldsymbol{b}_{2}\right),
\end{equation}
\begin{equation}
  \mathbf{c}_{i, r} =\sigma_{1}\left(\mathbf{W}_{1}\left[\mathbf{h}_{i}^{k-1} \oplus \mathbf{h}_{j}^{k-1} \oplus \mathbf{e}^{k-1}_r \oplus \mathbf{e}^{k-1}_{r_t}\right]+\mathbf{b}_{1}\right),
\end{equation}
where $ \mathcal{N}_r(i)$ denotes the immediate outgoing neighbors of node $ i $ under relation $ r $; $ \mathbf{W}_r^k $ is the transformation matrix for relation $ r $ for propagating messages; $ \sigma_1, \sigma_2 $ are Sigmoid function; $ \alpha_{i, r} $ is the attention weight of edge ($ i, r, j $); $ \phi(\mathbf{e}^{k-1}_r, \mathbf{h}_{j}^{k-1}) $ is a fusion operation to share hidden feature of nodes and relations. Inspired by \cite{vash2020compgcn}, we set the default fusion operation as subtraction $ \phi(\mathbf{e}, \mathbf{h}) = \mathbf{e} - \mathbf{h}$ to discriminate direction of relation.
In addition, to keep nodes and relations the same embedding space, relation embeddings are also transformed as follows:
\begin{equation}
  \mathbf{e}_r^{k} = \mathbf{W}_{rel}^{k} \mathbf{e}_r^{k-1}.
\end{equation} 

Inspired by CoMPILE \cite{mai2021compile}, we feed all node embeddings $\mathbf{H}^L$ of the last layer to a Gated Recurrent Unit (GRU) \cite{cho2014gru} to increase the expressive ability of network:
\begin{equation}
  \mathbf{H}^L = \operatorname{GRU}(\mathbf{H}^{L}).
\end{equation}

Finally, to obtain the representation $ \mathbf{h}_{\mathcal{G}} $ of subgraph $ \mathcal{G} $, we use an average readout function:
\begin{equation}
  \mathbf{h}_{\mathcal{G}} = \frac{1}{|{V}_{\mathcal{G}}|} \sum_{i \in {V}_{\mathcal{G}}} \mathbf{h}^L_i,
\end{equation}
where $ V_{\mathcal{G}} $ denotes the set of nodes in subgraph $ \mathcal{G} $.

\subsubsection{Neighboring Relational Path Module}

To solve the issue of sparse subgraph, we propose to explore neighboring relations to model \textit{neighboring relational paths}. This procedure can be seen in Figure \ref{fg:framework}. Specifically, a neighboring relational path is a relational sequence across the target nodes, i.e. $ p = (r_u, r_t, r_v) $, where $ r_u \in \mathcal{N}^{rel}(u)$ and $ r_v \in \mathcal{N}^{rel}(v) $ are relations around target nodes $ u $ and $ v $. We denote $ \mathcal{P}_{(u,v)} $ as the set of all neighboring relational paths across $ u $ and $ v $ in subgraph.

For each neighboring relational path $ p$, we first model it with Gated Recurrent Network (GRU) \cite{cho2014gru} as follows:
\begin{equation}
  \mathbf{p} = \operatorname{GRU}(p) = \operatorname{GRU}(\mathbf{e}_{r_u},\mathbf{e}_{r_t},\mathbf{e}_{r_v}).
\end{equation}
Then, we aggregate all path representations with attention to obtain the subgraph path representation $ \mathbf{p}_{\mathcal{G}} $: 
\begin{equation}
  \mathbf{p}_\mathcal{G} = \sum_{p \in \mathcal{P}} \beta_p \mathbf{p}
\end{equation}
\begin{equation}
  \beta_{p} = \frac{\text{exp}(\mathbf{p}^\top \mathbf{e}_{r_t})}{\sum_{p^\prime \in \mathcal{P}_{(u,v)}}\text{exp}(\mathbf{p}^{\prime \top} \mathbf{e}_{r_t})}.
\end{equation}

\subsubsection{Supervised Learning}
To organize above two modules in a unified framework, we combine the enclosing subgraph information $ \mathbf{h}_{\mathcal{G}} $ and neighboring relational path information $ \mathbf{p}_{\mathcal{G}} $ as the final representation of subgraph $ \mathbf{s}_{\mathcal{G}} $ :
\begin{equation}
  \mathbf{s}_{\mathcal{G}} = [\mathbf{h}_{\mathcal{G}} \oplus \mathbf{p}_{\mathcal{G}}],
\end{equation}
and assign score with embeddings of target triple $ (u, r_t, v) $:
\begin{equation}
  f(u, v_t, r) = \mathbf{W}_s[\mathbf{h}^L_u \oplus \mathbf{h}^L_v\oplus \mathbf{e}^L_{r_t} \oplus \mathbf{s}_{\mathcal{G}}],
\end{equation}
where $ \mathbf{h}^L_u, \mathbf{h}^L_u, \text{and } \mathbf{e}^L_{r_t}$ denote the embedding of target nodes $ u, v $ and target relation $ r $ in $ L $-th layer of GNN respectively. Finally, for supervised learning, we construct a margin-based loss function with equal negative triples by replacing heads or tails:
\begin{equation}\label{eq:sup}
  \mathcal{L}_{sup} = \sum_{(u, r_t, v)\in \mathcal{G}} \text{max}(0, f(u^\prime, r_t^\prime, v^\prime) - f(u, r_t, v) + \gamma),
\end{equation}
where$ (u, r_t, v) $ and $ (u^\prime, r_t^\prime, v^\prime) $ refer to positive and negative samples, and $ \gamma $ is the margin hyperparameter.

\subsection{MI Maximization in SNRI}

To avoid the subgraph neural network in SNRI over-emphasizing local structure, we further model neighboring relations in a global way by maximizing local-global (i.e. subgraph-graph) mutual information (MI), that is, we seek to enable neighboring relational features and paths to capture global information of entire KG.

To obtain global representation $ \mathbf{s}_{G}$ for $ G $, we use a readout function to summarize the obtained subgraph representations:
\begin{equation}
  \mathbf{s}_{G} = \frac{1}{N}\sum_{i=1}^{N}\mathbf{s}_{ \mathcal{G}_i},
\end{equation}
where $ N $ is the number of triples in knowledge graph $ G $; $ \mathcal{G}_i \in G$ is the subgraph of $ i $-th triple. 
Then, we utilize the Jensen-Shannon (JS) MI estimator \cite{Sun2021sugar} to maximize the estimated MI over subgraph and graph representations. Specifically, a discriminator $ \mathcal{D}(\mathbf{s}_{\mathcal{G}}, \mathbf{s}_{G}) $ is employed, which assign the probability score to subgraph-graph pair. Note that $ \mathcal{D} $ should be higher for subgraphs contained within the graph. Following DGI \cite{vel2019dgi}, we heuristically apply a bilinear function as the discriminator:
\begin{equation}
  \mathcal{D}(\mathbf{s}_{\mathcal{G}}, \mathbf{s}_{G}) = \sigma(\mathbf{s}_{\mathcal{G}}^\top \mathbf{W}_{MI} \mathbf{s}_{G}),
\end{equation}
where $ \sigma $ is the sigmoid function and $ \mathbf{W}_{MI} $ is a learnable scoring matrix. Since self-supervised MI mechanism is contrastive, negative graph $ G(\tilde{\mathbf{X}}, \tilde{\mathbf{A}}) $ is constructed by a corruption function $ \mathcal{C} $ : 
\begin{equation}
  \tilde{G}(\tilde{\mathbf{X}}, \mathbf{A}) \sim \mathcal{C}(G(\mathbf{X}, \mathbf{A})),
\end{equation}
where $ \mathbf{X} $ is the initial feature of nodes described in section 3.2, and $ \mathbf{A} $ is the adjacency matrix of $ G $. The corruption function $ \mathcal{C}(\cdot) $ preserves original structure but corrupts nodes feature by row-wise shuffling of $ \mathbf{X} $.

\begin{table*}[t]
  \centering
  \resizebox{\textwidth}{!}{
  \begin{tabular}{lrrrrrrrrrrrrrrrr}
    \toprule
    & \multicolumn{8}{c}{WN18RR} & \multicolumn{8}{c}{FB15k-237}\\
    \cmidrule(lr){2-9} \cmidrule(lr){10-17} 
    &\multicolumn{2}{c}{v1} & \multicolumn{2}{c}{v2} & \multicolumn{2}{c}{v3} & \multicolumn{2}{c}{v4} & \multicolumn{2}{c}{v1} & \multicolumn{2}{c}{v2} & \multicolumn{2}{c}{v3} & \multicolumn{2}{c}{v4} \\
     \cmidrule(lr){2-3} \cmidrule(lr){4-5} \cmidrule(lr){6-7} \cmidrule(lr){8-9} \cmidrule(lr){10-11} \cmidrule(lr){12-13} \cmidrule(lr){14-15} \cmidrule(lr){16-17}
    Method &AP &  H@10 & AP &  H@10 & AP &  H@10 & AP &  H@10 & AP &  H@10 & AP &  H@10 & AP &  H@10 & AP &  H@10 \\ 
    \midrule 
        Neural-LP & 86.02 &74.37  & 83.78 & 68.93 & 62.90 & 46.18 & 82.06 & 67.13 & 69.64 & 52.92  & 76.55 & 58.94 & 73.95 & 52.90 & 75.74 & 55.88  \\
    \text {DRUM}  & 86.02 & 74.37 & 84.05  & 68.93& 63.20 & 46.18 & 82.06 & 67.13 & 69.71 & 52.92  & 76.44 & 58.73 & 74.03 & 52.90 & 76.20 & 55.88  \\
    \text {RuleN} & 90.26 & 80.85 & 89.01  & 78.23& 76.46 & 53.39 & 85.75 & 71.59 & 75.24 & 49.76  & 88.70 & 77.82 & 91.24 & 87.69 & 91.79 & 85.60  \\
    \text {GraIL} & 94.32 & 82.45 & 94.18  & 78.68& 85.80 & 58.43 & 92.72 & 73.41 & 84.69 & 64.15  & 90.57 & 81.80 & 91.68 & 82.83 & 94.46 & 89.29   \\
    \text {CoMPILE} & 98.23 &83.60  & 99.56 &  79.82& 93.60 &  60.69& \textbf{99.80}  &75.49 & 85.50 &  67.64 & 91.68 & 82.98 & \textbf{93.12} & 84.67 & \textbf{94.90} & 87.44  \\
    \midrule 
    SNRI & \textbf{99.10} & \textbf{87.23} & \textbf{99.92} & \textbf{83.10} & \textbf{94.90} & \textbf{67.31} & 99.61 & \textbf{83.32} & \textbf{86.69} & \textbf{71.79} & \textbf{91.77} & \textbf{86.50} & 91.22 & \textbf{89.59} & 93.37 & \textbf{89.39} \\
    \bottomrule
  \end{tabular}}
  \caption{AUC-PR and Hits@10 results on the inductive benchmark datasets extracted from WN18RR and FB15k-237. We use AP and H@10 to denote AUC-PR and Hits@10, respectively. The best performance is highlighted.}
  \label{tb:main result}
\end{table*}

The MI objective for knowledge graph is realized by contrasting positive and negative subgraph-graph pairs: 
\begin{align}\label{eq:MI_loss}
  \mathcal{L}_{MI} &=\frac{1}{N+M}(\sum_{i=1}^{N} \mathbb{E}_{(\mathbf{X}, \mathbf{A})}\left[\log \mathcal{D}\left( \mathbf{s}_{\mathcal{G}_i}, \mathbf{s}_{G} \right)\right]  \nonumber \\
  &+ \sum_{j=1}^{M} \mathbb{E}_{(\tilde{\mathbf{X}}, {\mathbf{A}})}\left[\log \left(1-\mathcal{D}\left(\tilde{\mathbf{s}}_{\mathcal{G}_j}, \mathbf{s}_{G}\right)\right)\right]),
\end{align}
where $ N, M $ denote the number of positive and negative samples; $ \tilde{\mathbf{s}}_{\mathcal{G}} $ refers to the representation of negative subgraph sampled from $\tilde{G}$.

\begin{table}[t]
  \raggedright
  \resizebox{\columnwidth}{!}{
    \begin{tabular}{ccrrrrrr}
      \toprule
      & & \multicolumn{3}{c}{\text { WN18RR }} & \multicolumn{3}{c}{\text { FB15k-237 }} \\
      \cmidrule(lr){3-5} \cmidrule(lr){6-8} 
      & & \#R& \#N & \# T & \#R& \#N & \# T  \\
      \midrule
      \multirow{2}{*}{v1} 
      & train& 9 & 2746 & 6678 & 183 & 2000 & 5226 \\
      & test & 9 & 922 & 1991 & 146 & 1500 & 2404 \\
      \midrule
      \multirow{2}{*}{v2} 
      & train&  10 & 6954 & 18968 & 203 & 3000 & 12085 \\
      & test &  10 & 2923 & 4863 & 176 & 2000 & 5092 \\
      \midrule
      \multirow{2}{*}{v3} 
      & train& 11 & 12078 & 32150 & 218 & 4000 & 22394 \\
      & test & 11 & 5084 & 7470 & 187 & 3000 & 9137 \\
      \midrule
      \multirow{2}{*}{v4} 
      & train& 9 & 3861 & 9842 & 222 & 5000 & 33916 \\
      & test & 9 & 7208 & 15157 & 204 & 3500 & 14554 \\
      \bottomrule
    \end{tabular}}
    \caption{Statistics of inductive datasets. We use \#R, \#N, and \#T to denote the number of relations, nodes, and triples, respectively.} \label{tb:data}
\end{table}

\subsection{Joint Training Strategy}

The final learning objective of our work is defined as the combination of the supervised loss in Eq. \ref{eq:sup} and MI loss in Eq. \ref{eq:MI_loss}:
\begin{equation}
  \mathcal{L} = \mathcal{L}_{sup} + \lambda \mathcal{L}_{MI}, 
\end{equation}
where $ \lambda $ controls the contribution of the self-supervised MI mechanism. By this joint training strategy, our model is capable of modeling subgraph with complete relations while capturing neighboring relations aware of both local and global structural properties.

\section{Experiments}


\subsection{Experimental Configurations}

\paragraph{Datasets.} WN18RR \cite{dett2018wn18} and FB15k-237 \cite{tout2015fb} are common datasets used in transductive link prediction. For inductive link prediction task, we use the variants of WN18RR and FB15k-237 proposed by GraIL \cite{teru2020grail}, where entities in test set are not contained in train set and each dataset generate four versions datasets with increasing size. The statistics of the datasets is shown in Table \ref{tb:data}.

\paragraph{Evaluation protocol.} To compare fairly with the prior methods, we use the same evaluation protocol as \cite{teru2020grail}: AUC-PR for classification metrics and Hits@10 for ranking metrics. 
AUC-PR is an indicator for classification task by computing the area under the precision-recall curve. To compute AUC-PR, along with all positive triples in test set, we score an equal number of negative triples sampled by corrupting head or tail with a random entity. 
Hits@10 is the proportion of correct entities ranked in top 10 of candidate entities. For calculating Hits@10, we compare positive triples with sampled negative triples by assigning scores, to see whether the true triple can rank top 10.
Each result is obtained by averaging over 5 runs for accurate evaluation.

\paragraph{Hyperparameter settings.}  For subgraph extraction, we extract enclosing subgraph with 3 hops. In training process, we manually specify the hyperparameters as follows: learning rate to 0.001, dropout rate to 0.5, embedding dimension to 32. The margin $ \gamma $ in supervised loss function is set to 10, and coefficient $ \lambda $ in joint loss function is set to 5. The maximum number of training epochs is set to 30. We use Adam \cite{kinma2015adam} as optimizer to train our model. All experiments are implemented by PyTorch and run on NVIDIA RTX TITAN.

\paragraph{Baselines.} We compare our model to several state-of-the-art methods, including Neural-LP \cite{yang2015distmult}, DRUM \cite{sade2019drum}, GraIL \cite{teru2020grail}, and CoMPILE \cite{mai2021compile}. Neural-LP and DRUM are rule-based models learning logical rules and rule-confidence simultaneously in an end-to-end differentiable manner. GraIL and CoMPILE are graph-based models reasoning inductively by enclosing subgraph.

\begin{figure}[tp]
  \centering
  \includegraphics[scale=0.42]{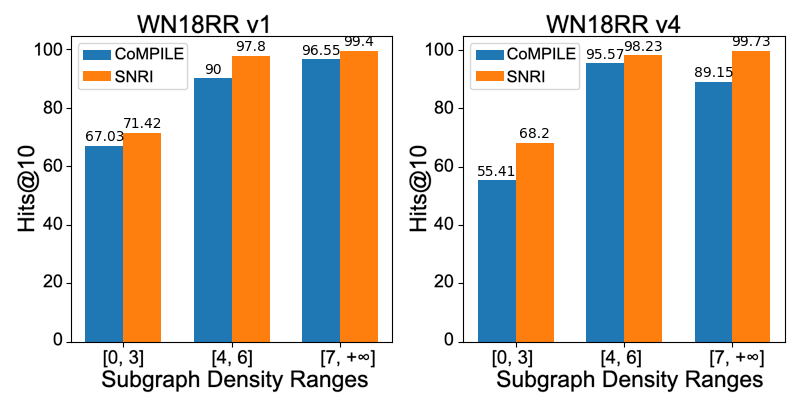}
  \caption{The performance comparison of SNRI and CoMPILE under different ranges of subgraph density on inductive WN18RR v1 and v4.} \label{fg:sparse result}
\end{figure}

\subsection{Main Results}

\paragraph{Comparison with baselines.} The results comparing with baseline models are shown in Table \ref{tb:main result}. The results show that our proposed model SNRI significantly outperforms baselines on the majority of datasets in terms of both AUC-PR and Hits@10 evaluation protocol, which demonstrates the effectiveness of our proposed model. Specifically, the average boosts of SNRI on WN18RR and FB15k-237 in Hits@10 reach up to 5.34\% and 3.64\% respectively compared with SOTA model CoMPILE, and it can be seen that the performance on WN18RR is more significant. This is because all previous works based on enclosing subgraph are hard to do reasoning when subgraph is sparse. As presented in Table \ref{tb:data} WN18RR has a lower ratio of \#T to \#E than FB15k-237, which means that subgraphs in WN18RR are more likely to be sparse, and thus scant of structure information for reasoning. In contrast, our model can deal with sparse subgraph powerfully by modeling neighboring relational features and neighboring relational paths to exploit complete neighboring relations sufficiently. But for FB15k-237, the improvement is less significant, which may be because subgraphs in FB15k-237 have very high density so that it is much easier for baselines to handle.

\paragraph{Effective modeling of sparse subgraph.}\label{sec:sparse} In this section, we tend to further verify that our proposed SNRI is capable of modeling complete neighboring relations to handle sparse subgraphs. We evaluate the ranking performance of CoMPILE and our proposed SNRI on subgraphs with different densities in WN18RR v1 and v4. Concretely, we divide subgraphs into three ranges according to the number of nodes in subgraph and then calculate Hits@10 of each range. As presented in Figure \ref{fg:sparse result}, SNRI performs better on WN18RR v1 and v4 across all ranges, especially for the range with low subgraph density. This result shows that SNRI possesses a better inductive ability for sparse subgraphs, and proves the necessity of capturing complete neighboring relations.

\subsection{Ablation Study}

\begin{table}
  \centering  
  \begin{tabular}{lrr}
    \toprule
    & \multicolumn{2}{c}{WN18RR} \\
    \cmidrule(lr){2-3}
    Method & v1 &v4\\
    \midrule
    SNRI w/o NRF & 86.96 & 82.26\\
    SNRI w/o NRP & 85.91 & 82.01\\
    SNRI w/o MI  & 84.84 & 82.43\\
    \midrule
    SNRI         & 87.23 & 83.32 \\
    \bottomrule
  \end{tabular}
  \caption{Ablation results of Hits@10 on inductive WN18RR v1 and v4.} \label{tb:ablation result}
\end{table}

In this section, we perform ablation study on WN18RR v1 and v4 to investigate the impact of each component in SNRI, namely, 1) neighboring relational feature (called \textbf{SNRI w/o NRF}), 2) neighboring relational paths (called \textbf{SNRI w/o NRP}), and 3) MI maximization (called \textbf{SNRI w/o MI}), by removing them respectively. Table \ref{tb:ablation result} shows the results of ablation studies. We can find that all variants of SNRI perform worse than the original SNRI, which demonstrates the effectiveness of each component.


\paragraph{\textbf{SNRI w/o NRF}.} After removing the neighboring relational feature, the Hits@10 value averagely reduces by 0.7\%. The reason may be that nodes feature with only positional information are less expressive, which cannot characterize the node effectively, and when nodes in subgraph are plentiful the positional feature is unstable and less robust. In contrast, complete neighboring relations are more effective and robust to characterize node features. 
\paragraph{\textbf{SNRI w/o NRP}.} From the result of SNRI w/o NRP, we can notice that performance drops a lot when neighboring relational paths are omitted. Associated with the result of section \ref{sec:sparse}, this result demonstrates neighboring relational paths are effective in handling sparse subgraphs. SNRI w/o NRF together with SNRI w/o NRP demonstrates the effectiveness of utilizing complete neighboring relations which are omitted by enclosing subgraph. 
\paragraph{\textbf{SNRI w/o MI}.} Additionally, removing MI maximization results in an average reduction of 1.7\%. This result implies global information is helpful to model neighboring relations better. We can observe that complete neighboring relations play a greater role in SNRI than MI maximization, but better performance can be obtained by considering complete neighboring relations and MI maximization simultaneously.

\subsection{Case Study}

\begin{table}
  \centering
  \resizebox{\columnwidth}{!}{
    \begin{tabular}{clr}
      \toprule
      Target Relations & Neighboring relational path & weight\\
      \midrule
      
      \multirow{2}{*}{\textit{related\_form}} &\textit{(related\_form, related\_form, \_also\_see)} & 0.51 \\
      &\textit{(related\_form,  related\_form, related\_form)} & 0.31 \\
      \midrule
      \multirow{2}{*}{\textit{adjoins}}&\textit{(country, adjoins, jurisdiction\_of\_office)} & 0.40 \\
      &\textit{(adjoin, adjoins, jurisdiction\_of\_office)} & 0.11 \\
      \midrule
      \multirow{2}{*}{\textit{dated\_participant}} & \textit{(people, dated\_participant, breakup\_participant)} & 0.99 \\
      &\textit{(award\_nominee, dated\_participant, dated\_participant)} & 0.01 \\
      \bottomrule
    \end{tabular}
  }
  \caption{Some neighboring relational paths with importance weight in inductive FB15k-237 dataset.} \label{tb:case result}
\end{table}
From WN18RR and FB15k-237, We select some target relations and then display the top 2 important neighboring relational paths in Table \ref{tb:case result}. The result shows that SNRI can learn correct neighboring relational paths and tend to assign a high score for the path with multiple relational types, indicating that SNRI prefers more informative neighboring relational paths to reason inductively. For example, considering target relation \textit{related\_form}, the path \textit{(related\_form, related\_form, \_also\_see)} gets a larger importance weight than \textit{(related\_form, related\_form, related\_form)} with single relational type.

\section{Conclusion}
In this paper, we propose a novel model called SNRI for inductive link prediction on knowledge graph, which can effectively exploit complete neighboring relations and learn global structure information. SNRI utilizes complete neighboring relations to characterize neighboring relational features of nodes in a more expressive manner, and then models neighboring relational path in a global way by MI maximization. The experiments on two benchmark datasets demonstrate our proposed SNRI significantly outperforms several existing state-of-the-art methods for the inductive link prediction task, and verify the effectiveness of modeling complete neighboring relations in a global way to characterize node features and reason on sparse subgraphs.

\section*{Acknowledgements}
The authors gratefully acknowledge the support of the National Natural Science Foundation of China (Grant No. 61876223, No. 61832004), Youth Innovation Promotion Association,Chinese Academy of Sciences (No.2020163), and International Cooperation and Exchanges NSFC (Grant No. 62061136006).









\bibliographystyle{named}
\bibliography{ijcai22}

\end{document}